%% file: main.tex
\documentclass{article}

\usepackage{microtype}
\usepackage{graphicx}
\usepackage{svg}
\usepackage{pgfplots}
\usetikzlibrary{patterns}
\usepackage{booktabs} 
\usepackage{subcaption}
\usepackage{longtable}
\usepackage{tablefootnote}

\usepackage{hyperref}

\usepackage[accepted]{icml2025}

\usepackage{amsmath}
\usepackage{amssymb}
\usepackage{mathtools}
\usepackage{amsthm}

\usepackage[capitalize,noabbrev]{cleveref}

\theoremstyle{plain}

\theoremstyle{definition}

\theoremstyle{remark}

\usepackage[textsize=tiny]{todonotes}

\icmltitlerunning{Minerva: A Programmable \textit{Memory Test} Benchmark for Language Models}

\begin{document}

\twocolumn[
\icmltitle{Minerva: A Programmable \textit{Memory Test} Benchmark for Language Models}



\icmlsetsymbol{equal}{*}

\begin{icmlauthorlist}
\icmlauthor{Menglin Xia}{equal,ms}
\icmlauthor{Victor R\"{u}hle}{ms}
\icmlauthor{Saravan Rajmohan}{ms}
\icmlauthor{Reza Shokri}{equal,nus}
\end{icmlauthorlist}

\icmlaffiliation{ms}{M365 Research, Microsoft}
\icmlaffiliation{nus}{National University of Singapore}

\icmlcorrespondingauthor{Menglin Xia}{mollyxia@microsoft.com}
\icmlcorrespondingauthor{Reza Shokri}{reza@comp.nus.edu.sg}

\icmlkeywords{LLM evaluation, memory test, context utilization, benchmark}

\vskip 0.3in
]



\printAffiliationsAndNotice{\icmlEqualContribution} 

\begin{abstract}
How effectively can LLM-based AI assistants utilize their memory (context) to perform various tasks? Traditional data benchmarks, which are often manually crafted, suffer from several limitations: they are static, susceptible to overfitting, difficult to interpret, and lack actionable insights--failing to pinpoint the specific capabilities a model lacks when it does not pass a test. In this paper, we present a framework for automatically generating a comprehensive set of tests to evaluate models' abilities to use their memory effectively. Our framework extends the range of capability tests beyond the commonly explored (passkey, key-value, needle in the haystack) search, a dominant focus in the literature. Specifically, we evaluate models on atomic tasks such as searching, recalling, editing, matching, comparing information in context memory, performing basic operations when inputs are structured into distinct blocks, and maintaining state while operating on memory, simulating real-world data. Additionally, we design composite tests to investigate the models' ability to perform more complex, integrated tasks. Our benchmark enables an interpretable, detailed assessment of memory capabilities of LLMs.

\end{abstract}

\input{sections/1_intro}

\input{sections/3_benchmark}
\input{sections/4_eval}

\input{sections/2_related_work}

\input{sections/5_conclusion}

\section*{Impact Statement}

This paper contributes to the advancement of Machine Learning by introducing a systematic and programmable evaluation framework for assessing the contextual processing capabilities of large language models. Our work provides insights into model strengths and limitations in handling various atomic and composite tasks, offering a structured way to analyze model behavior. These contributions can guide future research in improving model efficiency and reliability.

We do not foresee any direct ethical concerns or negative societal consequences arising from this work. Our evaluation methodology is designed to be model-agnostic and does not involve sensitive data or high-stakes applications.

\nocite{langley00}

\bibliography{references}
\bibliographystyle{icml2025}

\newpage
\appendix
\onecolumn
\input{sections/999_appendix_tasks}

\input{sections/999_appendix_hyperparams}

\end{document}

%% file: sections/1_intro.tex
\section{Introduction}

What capabilities should we expect from AI assistants? The AI assistants are provided with a (large) input context containing all the \textit{available} information that is potentially relevant to the user's request (e.g., all the prior emails, messages, and confirmed calendar events). This input, commonly referred to as the context (for the LLM), encapsulates what the AI assistant \textit{knows} about the world in which it is tasked to operate. This representation of the world, expressed in natural language, functions as the model's \textbf{memory}. In this paper, we address a fundamental question that is critical to improving AI assistants: \textit{What specific capabilities do large language models demonstrate in utilizing their memory?} 

One common approach to testing models is through data benchmarks. However, evaluating model capabilities using static data benchmarks--based on some user queries, their data, and expected outcomes--can be costly, imprecise, and lacks scalability. Additionally, performing tasks in realistic scenarios often requires multiple capabilities, making it challenging to identify which specific capability a model lacks when it fails a data benchmark. This limitation reduces the effectiveness of these benchmarks in designing better models. Moreover, blindly optimizing models to improve on such benchmarks risks overfitting, often rendering the data benchmarks obsolete over time.

To address some of these concerns, recently, there have been many attempts to test models using automatically generated benchmarks, but these efforts have primarily focused on evaluating basic search capabilities (e.g., passkey or key-value search) in long-contexts~\cite{kamradt2023, liu-etal-2024-lost, zhang-etal-2024-bench, anthropic2024b, wu2024longgenbench, li2024needlebench,hsieh2024ruler}. In this paper, we go beyond simple search tasks and introduce a framework to test a comprehensive range of memory-related capabilities in LLMs. Note that, we deliberately avoid conflating the \textit{memory usage} capabilities with the \textit{complex reasoning} abilities (e.g., complex mathematical or logical reasoning) in language models, as the latter is a separate skill that is currently the focus of extensive study~\cite{clark2018thinksolvedquestionanswering, cobbe2021gsm8k, hendrycksmath2021, suzgun2022challenging}.

\textit{What are memory-usage capabilities?} We define these as the abilities to retrieve relevant information, compose it for the instructed task, and recall key details when synthesizing the output. This process also involves creating associations between the instruction and the stored information, as well as among different parts of the memory itself. Without extracting these relationships, the memory remains flat and formless, rendering it less useful. Consequently, the model must be able to recognize differences, identify similarities, and take appropriate actions based on them. We design a series of \textit{atomic} tests aimed at evaluating each of these individual capabilities in isolation, to the extent that isolating such capabilities is possible.

To evaluate the more complex scenarios in memory usage, we construct composite tests that reflect real-world scenarios, where memory is divided into multiple compartments (i.e., information relevant to distinct contexts). The model is expected to recognize the boundaries of these compartments, trace the knowledge contained within them, and perform operations such as information retrieval, memory association, and other tasks while respecting these boundaries. The complexity increases further when there is \textit{interaction} between compartments. In such cases, information must flow across boundaries--for example, when stories about two parallel events converge at a particular moment, when interactions occur between different events, or when information known to certain entities in memory is shared with others. Examples include AI assistants managing calendar events, tracking financial transactions, or suggesting medical diagnoses. Handling these scenarios is highly challenging, yet it is essential for AI assistants to achieve practical and reliable performance. The core challenge is due to the fact that context memory, as provided to the model, flattens data from multiple parallel and potentially interrelated memory compartments. This requires the model to disentangle the content by leveraging the available labels and clues, and tracking the state of relevant information throughout the memory scanning, while performing the task.

Table~\ref{tab:memory_tests} presents an overview of the types of memory tests included in our benchmark. For each test, we use efficient parametric programs to generate randomized test cases. We run a comprehensive evaluation of several major open-source and black-box models (e.g., GPT-4(o), Cohere, Gemma, LLaMA, Mistral, Phi). Our experimental results show that while models perform relatively well on simple search tasks, they exhibit significant disparities across context utilization capabilities even at a context length of 4k tokens. This indicates that strong performance in basic retrieval does not necessarily translate to other context processing abilities. Our framework goes beyond search-based tests by incorporating atomic tests that pinpoint distinct capabilities, providing a more nuanced picture of the strengths and weaknesses of models in context processing. Moreover, composite tests, which combine multiple atomic capabilities, resulted in substantial performance drops for all models. These tests present the limitations of current models and provide valuable insights for guiding future model training and development.

The code and data will be available at \url{https://github.com/microsoft/minerva_memory_test}.

%% file: sections/3_benchmark.tex
\section{Benchmark for Memory Tests}

We define the entirety of context data available to a large language model (LLM) as its \textit{memory}. Users of an AI assistant leveraging the LLM can instruct the model to parse this memory and execute potentially complex retrieval tasks. Accordingly, we expect the AI assistant to demonstrate specific capabilities in memory utilization, including accurate retrieval, effective synthesis of relevant information, and adaptability to evolving context. These capabilities include:
\begin{itemize}
    \item \textbf{Information Retrieval and Localization}: The ability to efficiently locate, search for, and extract relevant information from the memory (i.e., input data) based on specific instructions or user queries.
    \item \textbf{Processing and Basic Reasoning}: The capability to perform modifications, computations, and logical operations on the input data, including identifying patterns, recognizing repetitions, and understanding relationships within the memory.
    \item \textbf{Content Transfer and Synthesis}: The ability to copy, rephrase, or generate synthesized output by integrating both original and modified elements from the input.
    \item \textbf{Structural Awareness and Organization}: The capacity to interpret the spatial, structural, or organizational layout of the memory, such as distinguishing labeled fragments of text, sets, lists, or hierarchical structures. 
\end{itemize}

In our benchmark, we focus on isolating the \textit{atomic} memory-related capabilities of LLMs. Success or failure in these tests provides a clear and interpretable assessment of the strengths and limitations of the models. We design multiple \textbf{atomic tests} to measure fundamental skills without interference from other factors. These tests are simple, targeted, and structured to assess specific abilities with clarity. Our benchmark includes the existing basic tests, notably the needle-in-the-haystack tests and its variations, but it goes beyond the search methods and includes a diverse set of atomic capabilities. In our benchmark, we focus on some fundamental capabilities: \textit{search, recall and edit, match and compare, spot the differences, compute on sets and lists, and stateful processing}. 

We also develop \textbf{composite tests} to evaluate how effectively models can perform more complex, integrated tasks. These tests assess the integration of multiple atomic capabilities to simulate real-world scenarios. By combining elements such as retrieval, reasoning, synthesis, and structural awareness, composite tests measure how well an AI assistant can coordinate different skills to execute complex operations. We provide two class of composite tests: processing data blocks, and composite-state tracking (theory of mind). Our objective here is to test the composition of various atomic operations and the ability to interpret segments of data (e.g., messages or paragraphs associated with a particular person or topic). These composite tests evaluate if the model can make sense of the ``spatial'' structure of the memory, and also keep track of its ``temporal'' changes (e.g., information that gets updated across many emails). This can become particularly challenging for the current architecture of major LLMs because the context has a flat structure. 

\input{tables/benchmark}
\input{figures/radar}
Table~\ref{tab:memory_tests} presents the list and description of representative tests in each category. Appendix~\ref{apd:prompt_example} presents the exact templates for all our tests. Our benchmark differs from traditional data benchmarks by allowing the generation of fresh, randomized test cases for each category. Each test acts as a \textit{programmable} script that measures the model’s capability while adjusting the hyperparameters that influence test difficulty. The programmable tests also enable us composing them easily, which is one of the key advantages of our benchmark. New categories, and new tests, can be easily added to this framework enabling a more diverse set of tests.

%% file: tables/benchmark.tex
\begin{table*}[t!]
\centering
\begin{tabular}{|p{5cm}|p{10cm}|}
%
\multicolumn{2}{l}{\textbf{Search}} \\ \hline
String search (binary test) & Is the string $x$ in memory? $x$ could be a word or a sequence.\\ \hline
Key-value search & What is the word or phrase that is paired with keyword $x$?\\ \hline
Batch search & For each of the keywords in the batch $x_1, x_2, \cdots, x_k$, perform the search and return the batch of corresponding responses.   \\ \hline
\multicolumn{2}{l}{\textbf{Recall and Edit}}    \\ \hline
Snapshot & Share a snapshot of the entire memory (as a verbatim copy). \\ \hline
Replace all & Share the entire memory after replacing all occurrences of $x$ with~$y$. \\ \hline
Overwrite positions & Share the entire memory after overwriting the words that are on particular positions (e.g., every $k^{\text{th}}$ word) with~$y$. \\ \hline
Functional updates & Update every $x$ with the output of a function $f(x)$.\\ \hline
\multicolumn{2}{l}{\textbf{Match and Compare}}    \\ \hline
Compare positions (binary test) & Does $x$ appear before $y$? \\ \hline
Find duplicates & Which word/string has duplicates in memory? \\ \hline
Count & How many times is $x$ repeated in memory? \\ \hline
Check association (binary test) & Check if both $x$ and $y$ are associated with the same tag in memory (assuming every term is associated with a tag).\\ \hline
\multicolumn{2}{l}{\textbf{Spot the Differences}}    \\ \hline
Compare two lists & Give two lists $X$ and $Y$ of the same length (e.g., same number of words), report the difference (i.e., $X - Y$). \\ \hline
Identify the odd group & Given multiple sets, identify which one is different (assuming $n-1$ identical sets with shuffled elements, and $1$ set with some differences). \\ \hline
Patch the difference & A sequence of words is repeated multiple times, and then there is a partial sequence. What is the next $k^{\text{th}}$ element in the sequence? \\ \hline
\multicolumn{2}{l}{\textbf{Compute on Sets and Lists}}    \\ \hline
Group membership & Given $k$ sets, identify which set includes $x$. \\ \hline
Group association & Check if $x$ and $y$ belong to the same set. \\ \hline
Iterate & Given $k$ lists, return the last element in each list.  \\ \hline
\multicolumn{2}{l}{\textbf{Stateful Processing}}    \\ \hline
Quantity & Keep track of the total quantity of items, based on a sequence of addition and subtraction operations (e.g., ``add~10, subtract~2, add~7, ...''). \\ \hline
Set & Keep track of the items in a set, based on a sequence of addition and removal operation (e.g., ``add~apple, pear; add~orange; remove~apple; add~lime, ...''). \\ \hline
\multicolumn{2}{l}{\textbf{Processing Data Blocks}}    \\ \hline
Search, recall, and edit & The input contains alternating labeled lists of elements (e.g., ``L1: a, b, c; L2: h, f, i; L1: d, z, k; ...''). For a given list label (e.g., L1 or L2) and a specified element within that list, return all the elements that appear after that specified element in the same list. \\ \hline
\multicolumn{2}{l}
{\textbf{Composite-State Tracking (Theory of Mind)}}    \\ \hline
State tracking across data blocks & Perform ``Stateful Processing'' for multiple agents, and report the final set state for each agent. The input provides a list of operations by $k$ agents over time, including both independent actions (add/remove) and interactive actions (swap) (e.g., ``Alice: add apple, pear, remove orange, add banana; Bob: add peach, berry, remove kiwi; Charley: add lime; Bob: remove peach, swap berry with Alice for banana; ...''). \\ \hline
\end{tabular}
\caption{List of memory tests. We divide the tests into different categories based on the core expected capability for passing the test. Most the initial tests are atomic, i.e., the expected capability cannot be broken down into other meaningful capabilities. The tests at the bottom of the list are composite tests and require the model to have multiple atomic capabilities at the same time in order to succeed.}
\label{tab:memory_tests}
\end{table*}

%% file: figures/radar.tex
\begin{figure*}[t!]
    \centering
    \begin{subfigure}[t]{0.45\textwidth}
        \includegraphics[width=0.9\textwidth]{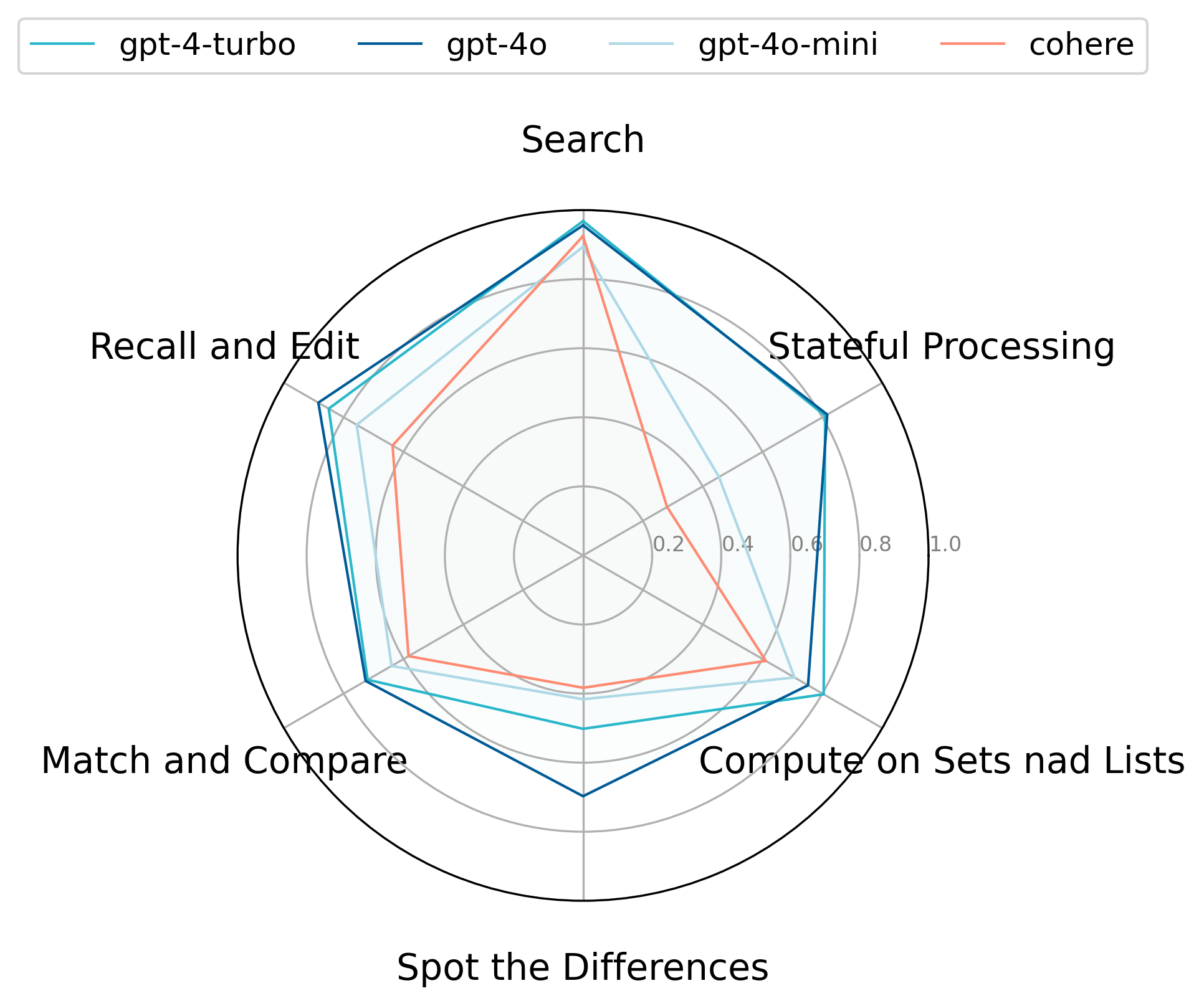}
        \caption{Performance of the black-box models.}
        \label{fig:first}
    \end{subfigure}
    \begin{subfigure}[t]{0.45\textwidth}
        \includegraphics[width=0.9\textwidth]{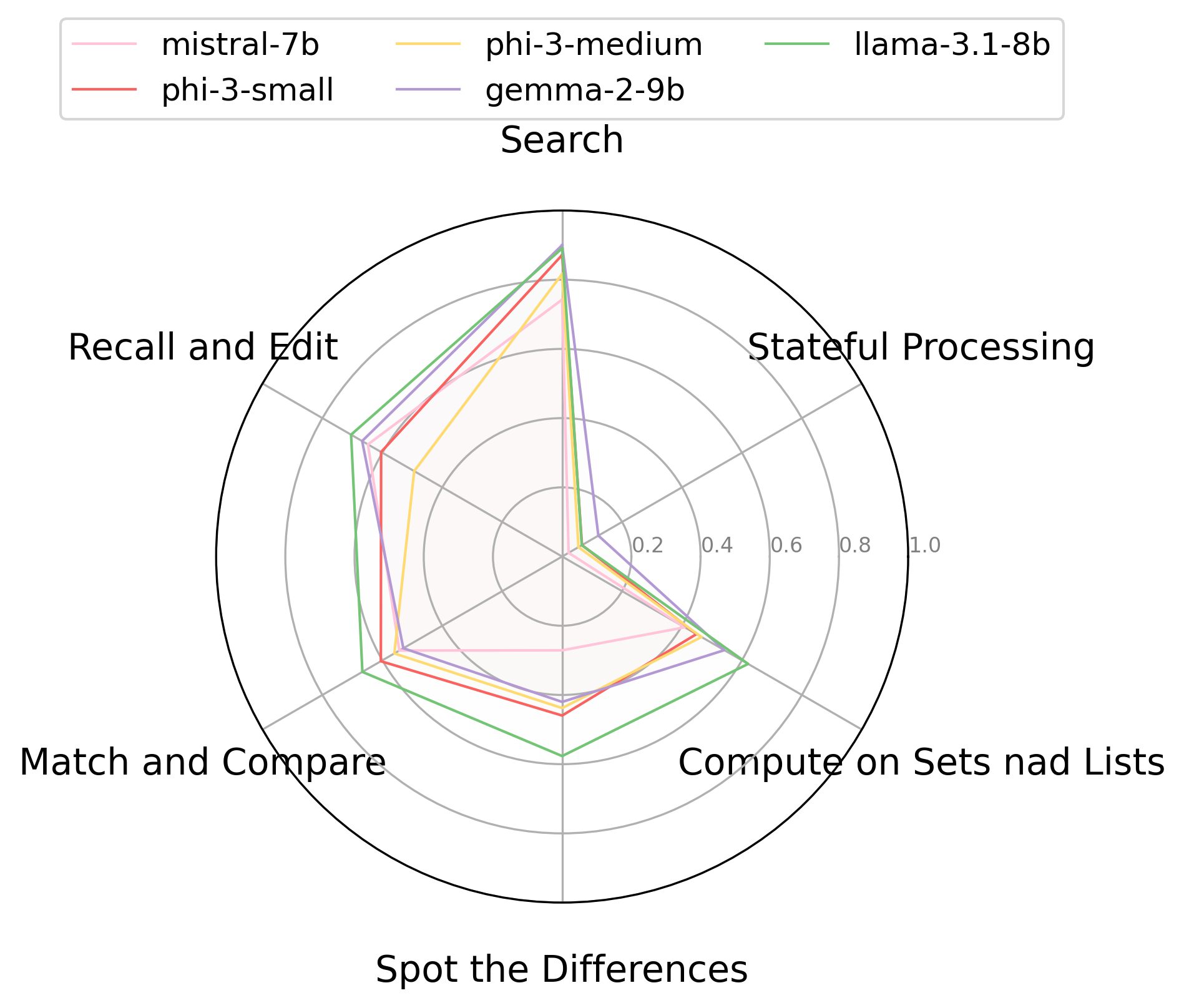}
        \caption{Performance of the open-source models.}
        \label{fig:second}
    \end{subfigure}
    \caption{Overall performance of nine models on a snapshot within 4k context length of Minerva.}
    \label{fig:radar}

\end{figure*}

%% file: sections/4_eval.tex
\section{Evaluation}

\subsection{Experimental Setup}
We use the proposed framework to evaluate nine widely used language models on a fixed snapshot of 1110 randomly generated test samples. For all tests, we fixed the context length to 4k tokens, except in the Stateful Processing category, where the context length depends on the number of operation steps. We set the number of steps as 200 for quantity state and 100 for set state, corresponding to an approximate context length of 1.5k tokens. For evaluation, we use exact match accuracy for binary tasks, ROUGE-L\citep{lin-2004-rouge} for tests that require sequence overlap measurement, and Jaccard similarity \citep{jaccard1901etude} for set overlap. Further details on the number of examples, hyperparameter configurations, and evaluation metrics for the tests are provided in Appendices \ref{apd:task_detail} and \ref{apd:eval}.

The evaluated models are divided into two groups: 

\textbf{Black-box models}: GPT-4-turbo, GPT-4o, GPT-4o-mini, and Cohere-command-rplus. 

\textbf{Open-source models}: Mistral-7b-instruct-v02, Phi-3-small-128k-instruct (7B), LLaMA-3.1-8b-instruct, Gemma-2-9b, and Phi-3-medium-128k-instruct (14B).

We set the max output token to 4096, temperature to 0, and top\_p to 1 for all model inference.

\subsection{Model Performance Overview}

Figure \ref {fig:radar} summarizes the overall performance of the evaluated models on the memory test snapshot within 4k context length. Notably, this context length is usually considered short for context utilization benchmarks, and many models are expected to perform perfectly at this length. However, our evaluation reveals significant disparities in performance across the capabilities, even within this manageable context length. Overall, the GPT-4-turbo/GPT-4o models show stronger all-around performance across the capabilities. In contrast, other models excel at the search task but struggle significantly in other areas, leading to a widening performance gap compared to stronger models. This is especially evident in the \textbf{Stateful Processing} tasks, where models exhibit steep performance drops. Even within the GPT-4(o) models, there were noticeable variations in performance across different tasks, despite them being the best-performing models. This suggests that strong performance in simple retrieval tasks does not imply effective context processing, highlighting that using NIAH-like tests alone for evaluating context utilization is not sufficient to capture the full spectrum of model capabilities. Our framework instead reveals significant variability in performance across distinct capability categories, offering a more nuanced understanding of model limitations.

The following sections analyze each test type in detail, highlighting key insights from the evaluations.

\subsection{Analysis on Atomic Tests}

\input{tables/search}
\paragraph{Search} All models performed relatively well on \textbf{Search} tasks, which is unsurprising given the 4k context length. However, even at this length, model performance varied significantly depending on the specific search type (see Table \ref{tab:search}). For example, in the binary \textit{String Search} task, models handled individual word searches well but struggled with subsequence searches, where queries consisted of multi-word sequences. The performance drop can be attributed to two factors: (1) length of query affects the difficulty of precise memory access; (2) negative samples are created by replacing a single word in present subsequences, making absent longer subsequence more distracting.

\input{figures/ablation_seq_search}

Figure \ref{fig:seq_search} further analyzes subsequence search performance for GPT-4o, Mistral, and Phi-3-medium. These models exhibit distinct error patterns as the length of the subsequence increases: GPT-4o has no false negative errors (it never misses a present subsequence) but makes more false positive errors as the subsequence length grows, suggesting it overestimates presence in more ambiguous cases.
Mistral also makes no false negative errors but exhibits a decreasing false positive rate, implying it struggles more with shorter distractors. Phi-3-medium, in contrast, makes few false positive errors (rarely identifies an absent sequence as present), but struggles more with false negatives, indicating a general tendency to deny presence. These differing patterns suggest that the models may employ different search strategies, affecting their susceptibility to different types of errors.

For \textit{Batch Search} and \textit{Key-Value Search} tasks (analogous to multi-NIAH and NIAH, respectively), models like Mistral, Phi-3, and Cohere show a notable performance drop, revealing their limitations in handling multiple memory accesses effectively.

\input{figures/recall}
\paragraph{Recall and Edit} 
\input{tables/ablation_gibberish}

Figure \ref{fig:recall} presents the results for the \textbf{Recall and Edit} tasks. While models performed well on basic recall (\textit{Snapshot}), their performance dropped sharply when tasked with making regular edits. A closer analysis of the generated outputs reveals that models struggled with maintaining coherence during edits, often getting trapped in repetitive word loops. For the \textit{Functional Update} task, we deliberately selected simple numerical updates, such as ``Subtract 1 from every number," to ensure the edits were within the models' capabilities. Nevertheless, when comparing performance on \textit{Snapshot (with numbers)} to \textit{Functional Updates}, all models exhibited a steep decline, especially for smaller ones. Analysis of generated outputs revealed that these models frequently deviated from instructions over longer sequences, suggesting difficulties in maintaining consistent rule applications over extended contexts.

Additionally, we conducted a separate ablation study on \textit{Snapshot} and \textit{String Search}. In this study, we replaced meaningful words in the context with gibberish tokens consisting of randomly generated alphabetical characters. As shown in Table \ref{tab:ablation_gibberish}, performance remained largely unchanged, suggesting that semantic meaning was not a significant distractor in these tasks.

\input{figures/compare}

\input{tables/group}

\paragraph{Match and Compare}
 As shown in Figure \ref{fig:match}, model performance in the \textbf{Match and Compare} tasks was relatively consistent across different model sizes. Given that counting is a well-known weakness in LLMs, it is unsurprising that all models struggled significantly with the counting task, though GPT models performed slightly better than others. However, models generally succeeded in identifying the duplicates (in \textit{Find duplicates}), and primarily struggled with the counting aspect, which requires tracking and updating an integer state, a skill that is more similar to stateful processing. This suggests that relying solely on counting-based tests \cite{song2024countingstars} could overly bias the evaluation and fail to capture broader model capabilities. The results also indicate that models exhibit some ability to recognize relative positions and group associations, but their accuracy remains limited (ranging between 0.6-0.8). A closer examination of model generations reveals an overwhelming tendency for the models to produce false positive errors -- models often answer “yes” when the correct answer is “no”, while making very few false negative errors. This means that when the relationship is correct, the models can more reliably identify it. This may stem from a combination of their inherent inclination to agree and the difficulty in recognizing relative comparisons and associations.

\input{figures/difference}

\paragraph{Spot the Differences}
As shown in Figure \ref{fig:difference}, performance across all models are poor on \textit{Compare Two Lists}, suggesting inherent difficulties in cross-referencing information across long contexts, even for larger models.  GPT-4o and the LLaMA model significantly outperform the others in the \textit{Identify the Odd Group} task, highlighting a general weakness in detecting contextual differences by the other models. However, an 8B LLaMA model outperforms both equivalently-sized models and even GPT-4 in this task, suggesting that model size alone was not the determining factor. This indicates that architectural differences, training objectives, or specific inductive biases may contribute to improved performance in comparative memory utilization.

\paragraph{Compute on Sets and Lists}
The tasks in this category require models to recognize and process group structures within the context, and performance gradually declines as the complexity of the task increases (see Table \ref{tab:lists}). For instance, in comparing the \textit{Group Membership} task with the \textit{String Search} task, where the former requires identifying which list a word belongs to rather than simply determining its presence, the performance of open-source models drops considerably. Similarly, in comparing the \textit{Group Association} task with the \textit{Group Membership} task, where the former requires determining whether two words belong to the same group, all models exhibit a noticeable decline in performance. The decline becomes even more pronounced when comparing the \textit{ Group Association (alternating)} variant of the task to the standard \textit{Group Association} task. Here, the context involves alternating repeated groups rather than simple group structures, which further challenges the models' abilities to handle partitioned contexts effectively.

An interesting observation was found during the \textit{Iterate} task. In an ablation study, we modified the task to require returning the first words in each list instead of the last words (making it more similar to the \textit{Batch Search} task). The performance sharply declines when models are asked to return the last words, despite their strong information-fetching capabilities. This suggests that, while the models can retrieve information effectively, they struggle to accurately recognize and process partitions within the context.

\input{tables/state}
\paragraph{Stateful Processing}

\input{figures/ablation_state_step}

Table \ref{tab:state} presents the results for the \textbf{Stateful Processing} tasks, where performance gaps among models are the most pronounced. The GPT-4(o) models perform well on integer state tracking, while most other models struggle (near zero accuracy). For set state tracking, larger models generally perform better.

We conducted an ablation study to examine how the number of operation steps influences performance of five selected models (Fig. \ref{fig:ablation_state_step}). For quantity state tracking, GPT-4(o) models perform well within fewer than 200 steps but experience a sharp decline in accuracy beyond this threshold. For set state tracking, the performance decline is more gradual. The differences in performance drop between the two tasks can be attributed to the nature of the two tasks. While tracking an integer state might seem simpler than tracking a set, it actually requires the model to maintain and apply every operation sequentially to compute the final value. In contrast, for set state, the fixed size of the set makes more recent operations more relevant to the final state, reducing the need for exhaustive step-by-step tracking. Nevertheless, even in this scenario, all models show a clear inability to handle longer or more complex operation sequences effectively. Interestingly, GPT-4 model outperformed GPT-4o at this task, suggesting potential optimization trade-offs may have affected its ability to manage set-based updates. 

Overall, while larger models like GPT-4(o) exhibit some ability to track state over time, their effectiveness rapidly deteriorates as task complexity increases. Smaller models, in particular, struggle to track operations over time, pointing to significant gaps in their ability to manage and process sequential dependencies critical for state tracking tasks.

\subsection{Results on Composite Tests}

\input{tables/composition}

The composite tests significantly challenge the models by combining multiple atomic capabilities into a single test. In the \textit{Processing Data Blocks} task, the context is fixed at 4k tokens, while for the \textit{Theory of Mind} task, the number of operation steps is set to 100. As shown in Table \ref{tab:comp}, model performance on both tasks are generally low, showing a broad inability to handle the more complex scenarios. Performance across all models drop substantially on composite tasks compared to their performance on individual capability tasks, such as search, recall, and group processing. 

Interestingly, some smaller models, like Mistral and Phi-3-small, exhibit slightly better performance on the \textit{Theory of Mind} task than on the set state tracking task. This anomaly likely stems from their already weak state tracking ability, which limits their performance across both tasks. Additionally, these models tend to generate longer answers in the set state task which reduces the set overlap.

Notably, even the most capable models, such as GPT-4-turbo and GPT-4o, struggle, showing that scaling model size alone is not enough for solving these composite tasks. Additionally, the variation in performance among smaller models suggests that their limitations stem not only from size but also from underlying architectural or training differences. This indicates that smaller models require more targeted care to bridge the gap in effective memory use.

\subsection{Extending the benchmark to other configurations}

Our benchmark is fully programmable and supports flexible experimentation across a wide range of configurations, including varying context lengths, evaluation criteria, and prompt phrasing. In this section, we illustrate how the benchmark can be adapted beyond the default setup used in the main paper. These examples highlight its versatility in probing model behavior under diverse conditions.

\paragraph{Context length}
In the main experiments, we fixed the context length to 4K tokens to emphasize that models already exhibit notable failures at this moderate length across many memory tasks. However, the benchmark is scalable to longer contexts. Table~\ref{tab:context_length_combined} presents additional results for two representative tasks \textit{Functional updates} and \textit{Counting} evaluated at various context lengths up to 16K tokens.

In both cases, we observe that model performance begins to degrade significantly well before reaching what is typically considered a ``long" context window. These failures reveal underlying limitations in how models manage long-range memory beyond simple retrieval. In contrast, models tend to maintain strong performance on retrieval-style tasks (e.g., \textit{String search}) even at extended lengths, making such tasks less effective at distinguishing model capabilities. Additional results and task examples are provided in Appendix~\ref{apd:more_examples}.

\begin{table*}[h]
\centering
\begin{subtable}[t]{0.48\textwidth}
\centering
\resizebox{0.8\columnwidth}{!}{
\begin{tabular}{lccccc}
\toprule
\textbf{Model} & 500 & 1K & 2K & 4K & 8K \\
\midrule
gpt-4o & 1.00 & 1.00 & 0.99 & 0.93 & 0.59 \\
gpt-4o-mini & 0.69 & 0.66 & 0.42 & 0.24 & 0.10 \\
phi-3-small & 0.49 & 0.45 & 0.21 & 0.07 & 0.03 \\
\bottomrule
\end{tabular}}
\caption{Functional updates (ROUGE-L)}
\end{subtable}
\hfill
\begin{subtable}[t]{0.48\textwidth}
\centering
\resizebox{0.85\columnwidth}{!}{
\begin{tabular}{lccccc}
\toprule
\textbf{Model} & 1K & 2K & 4K & 8K & 16K \\
\midrule
gpt-4-turbo & 0.52 & 0.44 & 0.40 & 0.32 & 0.28 \\
cohere & 0.36 & 0.24 & 0.20 & 0.20 & 0.12 \\
phi-3-medium & 0.20 & 0.16 & 0.12 & 0.08 & 0.04 \\
\bottomrule
\end{tabular}}
\caption{Counting (Exact match)}
\end{subtable}
\caption{Performance across context lengths on two representative tasks.}
\label{tab:context_length_combined}
\end{table*}

\paragraph{Prompt Variation}

We also investigated the model performance sensitivity to minor variations in prompt phrasing. Specifically, we tested different phrasings of task instructions while keeping the underlying task logic unchanged. For instance, in the \textit{String search} task, we compared “\textit{Given the context, determine if XXX is present} (Var 1)” versus “\textit{Is XXX present in the context?} (Var 2)”. Similarly, in the \textit{Group association} task, we tested “\textit{Determine if word ‘AAA’ and word ‘BBB’ are in the same list} (Var 1)” versus “\textit{Check if the words ‘AAA’ and ‘BBB’ belong to the same list} (Var 2)”.

As shown in Table~\ref{tab:prompt_variation_combined}, performance differences between prompt variants are generally small, suggesting that instruction interpretation is not the primary bottleneck in these tasks, rather, the main challenge lies in actually executing the task correctly. Nevertheless, we recognize that more intensive prompt engineering could potentially affect model performance. Given these findings, we standardized the prompts to a single, simple version (as shown in Appendix \ref{apd:prompt_example}) for all experiments in this paper to ensure consistency and comparability across models. However, because the benchmark is programmable, researchers can easily swap in alternate prompts to explore additional prompt settings. Appendix~\ref{apd:more_examples} provides additional examples on more tests for prompt variations.

\begin{table*}[h]
\centering
\begin{subtable}[t]{0.48\textwidth}
\centering
\resizebox{0.87\columnwidth}{!}{
\begin{tabular}{lcccc}
\toprule
\textbf{Model} & \textbf{Var 1} & \textbf{CI$_{95\%}$} & \textbf{Var 2} & \textbf{CI$_{95\%}$} \\
\midrule
gpt-4o        & 1.00 & (0.93, 1.00)  & 1.00 & (0.93, 1.00)   \\
gpt-4o-mini   & 0.98 & (0.90, 1.00)  & 0.98 & (0.90, 1.00) \\
phi-3-medium  & 1.00 & (0.93, 1.00) & 1.00 & (0.93, 1.00)  \\
\bottomrule
\end{tabular}}
\caption{\textbf{String search} (Exact match)}
\end{subtable}
\hfill
\begin{subtable}[t]{0.48\textwidth}
\centering
\resizebox{0.85\columnwidth}{!}{
\begin{tabular}{lcccc}
\toprule
\textbf{Model} & \textbf{Var 1} & \textbf{CI$_{95\%}$} & \textbf{Var 2} & \textbf{CI$_{95\%}$} \\
\midrule
gpt-4o        & 0.65 & (0.50, 0.78) & 0.63 & (0.47, 0.76)  \\
cohere        & 0.70 & (0.55, 0.82) & 0.75 & (0.60, 0.86) \\
phi-3-small   & 0.55 & (0.40, 0.69) & 0.55 & (0.40, 0.69) \\
\bottomrule
\end{tabular}}

\caption{\textbf{Group association} (Exact match)}
\end{subtable}
\caption{Prompt variation performance on the String search task (with 50 samples) and the Group association task (with 40 samples). Variation 1 and Variation 2 differ slightly in phrasing but preserve task intent.}
\label{tab:prompt_variation_combined}
\end{table*}

%% file: tables/search.tex
\begin{table}[h]
    \centering
    \resizebox{\columnwidth}{!}{%
    \begin{tabular}{l|llll}
        \toprule
        \textbf{Models} & \textbf{Word} & \textbf{Subsequence} &\textbf{ Key-value} &\textbf{ Batch} \\ \midrule
gpt-4-turbo & 0.94 & 0.94 (-0.00) & 1.00 & 1.00 (-0.00) \\
gpt-4o & 1.00 & 0.82 \textcolor{red}{(-0.18)} & 1.00 & 1.00 (-0.00) \\
gpt-4o-mini & 0.98 & 0.64 \textcolor{red}{(-0.34)} & 1.00 & 0.96 \textcolor{red}{(-0.04)}\\
cohere-command-rplus & 1.00 & 0.85 \textcolor{red}{(-0.15)} & 0.98 & 0.87 \textcolor{red}{(-0.11)}\\
mistral-7b  & 0.78 & 0.80 \textcolor{green}{
(+0.02)} & 0.92 & 0.47 \textcolor{red}{(-0.45)} \\
phi-3-small & 0.94 & 0.84 \textcolor{red}{(-0.10)} & 0.94 & 0.77 
 \textcolor{red}{(-0.17)}\\
phi-3-medium & 1.00 & 0.55 \textcolor{red}{(-0.45)} & 1.00 & 0.72 \textcolor{red}{(-0.28)}\\
gemma-2-9b & 1.00 & 0.60 \textcolor{red}{(-0.40)} & 1.00 & 1.00 (-0.00) \\
llama-3.1-8b & 1.00 & 0.57 \textcolor{red}{(-0.43)}  & 1.00 & 0.99 \textcolor{red}{(-0.01)} \\ 
        \bottomrule
    \end{tabular}
    }
    \caption{\textbf{Results for the Search tasks.} The four columns represent: String Search (with word), String Search(with subsequence), Key-value Search, and Batch Search. Numbers in parentheses indicate comparative performance differences between String Search (with subsequence vs. word) and Batch Search vs. Key-Value Search.}
    \label{tab:search}
\end{table}

%% file: figures/ablation_seq_search.tex

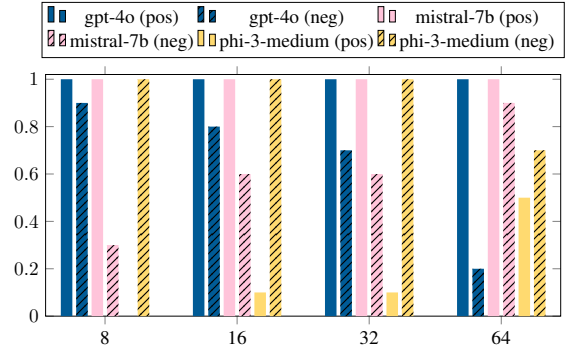
\begin{figure}
    \centering
\resizebox{0.9\columnwidth}{!}{
    \begin{tikzpicture}
        \begin{axis}[
            ybar,
            bar width=6pt,
            symbolic x coords={8, 16, 32, 64},
            xtick=data,
            ymin=0, ymax=1.02,
            legend columns=3,
            legend style={at={(0.5,1.3)}, anchor=north, draw=black},
            enlarge x limits=0.15,
            width=11cm, height=6cm
        ]
        
        \addplot[fill={rgb,255:red,0;green,91;blue,150}, draw=none] coordinates {(8,1.000) (16,1.000) (32,1.000) (64,1.000)};
        \addlegendentry{gpt-4o (pos)}
        \addplot[fill={rgb,255:red,0;green,91;blue,150}, postaction={
        pattern=north east lines
    }, draw=none] coordinates {(8,0.900) (16,0.800) (32,0.700) (64,0.200)};
        \addlegendentry{gpt-4o (neg)}
        
        \addplot[fill={rgb,255:red,255;green,196;blue,218}, draw=none] coordinates {(8,1.000) (16,1.000) (32,1.000) (64,1.000)};
        \addlegendentry{mistral-7b (pos)}
        \addplot[fill={rgb,255:red,255;green,196;blue,218}, postaction={
        pattern=north east lines
    }, draw=none] coordinates {(8,0.300) (16,0.600) (32,0.600) (64,0.900)};
        \addlegendentry{mistral-7b (neg)}
        
        \addplot[fill={rgb,255:red,255;green,218;blue,112}, draw=none] coordinates {(8,0.000) (16,0.100) (32,0.100) (64,0.500)};
        \addlegendentry{phi-3-medium (pos)}
        \addplot[fill={rgb,255:red,255;green,218;blue,112}, postaction={
        pattern=north east lines
    }, draw=none] coordinates {(8,1.000) (16,1.000) (32,1.000) (64,0.700)};
        \addlegendentry{phi-3-medium (neg)}
        
        \end{axis}
    \end{tikzpicture}}
    \caption{Analysis on \textit{String Search (with subsequence)} across increasing subsequence lengths. This figure examines the behavior of models on \textbf{pos}itive samples (where the subsequence is present) and \textbf{neg}ative samples (where the subsequence is absent).}
    \label{fig:seq_search}
\end{figure}

%% file: figures/recall.tex

\begin{figure}[h]
    \centering
    \begin{subfigure}{0.49\columnwidth}
        \resizebox{\textwidth}{!}{\input{figures/recall_black}}
    \end{subfigure}
    \begin{subfigure}{0.49\columnwidth}
        \resizebox{\textwidth}{!}{\input{figures/recall_white}}
    \end{subfigure}
    \caption{Results for the \textbf{Recall and Edit} tasks.}
    \label{fig:recall}
\end{figure}
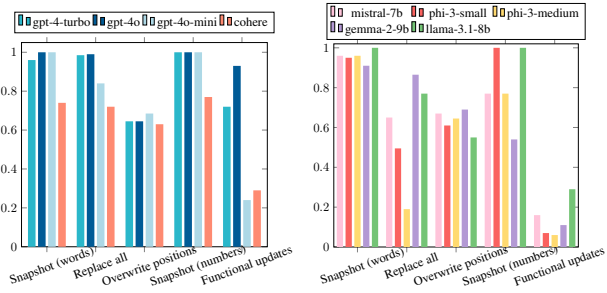

%% file: figures/recall_black.tex
\begin{tikzpicture}
        \begin{axis}[
            ybar,
            bar width=6pt,
            symbolic x coords={Snapshot (words), Replace all, Overwrite positions, Snapshot (numbers), Functional updates},
            xtick=data,
            ymin=0, ymax=1.05,
            legend columns=4,
            legend style={at={(0.5,1.15)}, anchor=north, draw=black},
            enlarge x limits=0.13,
            xticklabel style={rotate=20, anchor=center, yshift=-12pt}
        ]
        
        \addplot[fill={rgb,255:red,42;green,183;blue,202}, draw=none] coordinates {(Snapshot (words),0.96) (Replace all,0.985) (Overwrite positions,0.645) (Snapshot (numbers),1.00) (Functional updates,0.72)};
        \addlegendentry{gpt-4-turbo}
        
        \addplot[fill={rgb,255:red,0;green,91;blue,150}, draw=none] coordinates {(Snapshot (words),1.00) (Replace all,0.99) (Overwrite positions,0.645) (Snapshot (numbers),1.00) (Functional updates,0.93)};
        \addlegendentry{gpt-4o}
        
        \addplot[fill={rgb,255:red,173;green,216;blue,230}, draw=none] coordinates {(Snapshot (words),1.00) (Replace all,0.84) (Overwrite positions,0.685) (Snapshot (numbers),1.00) (Functional updates,0.24)};
        \addlegendentry{gpt-4o-mini}
        
        \addplot[fill={rgb,255:red,254;green,138;blue,113}, draw=none] coordinates {(Snapshot (words),0.74) (Replace all,0.72) (Overwrite positions,0.63) (Snapshot (numbers),0.77) (Functional updates,0.29)};
        \addlegendentry{cohere}
        
        \end{axis}
\end{tikzpicture}

%% file: figures/recall_white.tex
    \begin{tikzpicture}
        \begin{axis}[
            ybar,
            bar width=5pt,
            symbolic x coords={Snapshot (words), Replace all, Overwrite positions, Snapshot (numbers), Functional updates},
            xtick=data,
            ymin=0, ymax=1.02,
            legend columns=3,
            legend style={at={(0.5,1.20)}, anchor=north, draw=black},
            enlarge x limits=0.12,
            xticklabel style={rotate=20, anchor=center, yshift=-12pt}
        ]
        
        \addplot[fill={rgb,255:red,255;green,196;blue,218}, draw=none] coordinates {(Snapshot (words),0.96) (Replace all,0.65) (Overwrite positions,0.67) (Snapshot (numbers),0.77) (Functional updates,0.16)};
        \addlegendentry{mistral-7b}
        
        \addplot[fill={rgb,255:red,250;green,98;blue,95}, draw=none] coordinates {(Snapshot (words),0.95) (Replace all,0.495) (Overwrite positions,0.61) (Snapshot (numbers),1.00) (Functional updates,0.07)};
        \addlegendentry{phi-3-small}
        
        \addplot[fill={rgb,255:red,255;green,218;blue,112}, draw=none] coordinates {(Snapshot (words),0.96) (Replace all,0.19) (Overwrite positions,0.645) (Snapshot (numbers),0.77) (Functional updates,0.06)};
        \addlegendentry{phi-3-medium}
        
        \addplot[fill={rgb,255:red,179;green,153;blue,212}, draw=none] coordinates {(Snapshot (words),0.91) (Replace all,0.865) (Overwrite positions,0.69) (Snapshot (numbers),0.54) (Functional updates,0.11)};
        \addlegendentry{gemma-2-9b}
        
        \addplot[fill={rgb,255:red,116;green,196;blue,118}, draw=none] coordinates {(Snapshot (words),1.00) (Replace all,0.77) (Overwrite positions,0.55) (Snapshot (numbers),1.00) (Functional updates,0.29)};
        \addlegendentry{llama-3.1-8b}
        
        \end{axis}
\end{tikzpicture}

%% file: tables/ablation_gibberish.tex
\begin{table}[!h]
    \centering
        \resizebox{0.8\columnwidth}{!}{%
    \begin{tabular}{lllll}
    \toprule
        \textbf{Model} & \textbf{String Search (word)} & \textbf{Snapshot} \\ \hline
gpt-4-turbo    & 1.00 \textcolor{green}{(0.06)} & 1.00 \textcolor{green}{(0.04)} \\ 
gpt-4o         & 1.00 (0.00)                   & 1.00 (0.00)                   \\ 
gpt-4o-mini    & 0.94 \textcolor{red}{(-0.04)}  & 1.00 (0.00)                   \\ 
cohere         & 1.00 (0.00)                   & 1.00 \textcolor{green}{(0.26)} \\ 
mistral-7b     & 1.00 \textcolor{green}{(0.22)} & 0.96 (0.00)                   \\ 
phi-3-small    & 1.00 \textcolor{green}{(0.06)} & 0.99 \textcolor{green}{(0.04)} \\ 
phi-3-medium   & 0.98 \textcolor{red}{(-0.02)}  & 0.87 \textcolor{red}{(-0.09)}  \\ 
gemma-2-9b     & 0.96 \textcolor{red}{(-0.04)}  & 0.96 \textcolor{green}{(0.05)} \\ 
llama-3.1-8b   & 0.98 \textcolor{red}{(-0.02)}  & 1.00 (0.00)                   \\
\bottomrule
    \end{tabular}
    }
    \caption{Ablation study with gibberish context.}
    \label{tab:ablation_gibberish}
\end{table}

%% file: figures/compare.tex

\begin{figure}[h]
    \centering
    \begin{subfigure}{0.49\columnwidth}
        \resizebox{\textwidth}{!}{\input{figures/compare_black}}
    \end{subfigure}
    \begin{subfigure}{0.49\columnwidth}
        \resizebox{\textwidth}{!}{\input{figures/compare_white}}
    \end{subfigure}
    \caption{Results for the \textbf{Match and Compare} tasks.}
    \label{fig:match}
\end{figure}
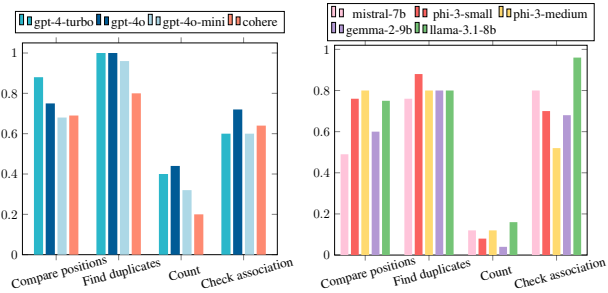

%% file: figures/compare_black.tex
\begin{tikzpicture}
        \begin{axis}[
            ybar,
            bar width=7pt,
            symbolic x coords={Compare positions, Find duplicates, Count, Check association},
            xtick=data,
            ymin=0, ymax=1.05,
            legend columns=4,
            legend style={at={(0.5,1.15)}, anchor=north, draw=black},
            enlarge x limits=0.18,
            xticklabel style={rotate=15, anchor=center, yshift=-10pt}
        ]
        
        \addplot[fill={rgb,255:red,42;green,183;blue,202}, draw=none] coordinates {(Compare positions,0.88) (Find duplicates,1.00) (Count,0.40) (Check association,0.60)};
        \addlegendentry{gpt-4-turbo}
        
        \addplot[fill={rgb,255:red,0;green,91;blue,150}, draw=none] coordinates {(Compare positions,0.75) (Find duplicates,1.00) (Count,0.44) (Check association,0.72)};
        \addlegendentry{gpt-4o}
        
        \addplot[fill={rgb,255:red,173;green,216;blue,230}, draw=none] coordinates {(Compare positions,0.68) (Find duplicates,0.96) (Count,0.32) (Check association,0.60)};
        \addlegendentry{gpt-4o-mini}
        
        \addplot[fill={rgb,255:red,254;green,138;blue,113}, draw=none] coordinates {(Compare positions,0.69) (Find duplicates,0.80) (Count,0.20) (Check association,0.64)};
        \addlegendentry{cohere}
        
        \end{axis}
\end{tikzpicture}

%% file: figures/compare_white.tex
\begin{tikzpicture}
        \begin{axis}[
            ybar,
            bar width=6pt,
            symbolic x coords={Compare positions, Find duplicates, Count, Check association},
            xtick=data,
            ymin=0, ymax=1.02,
            legend columns=3,
            legend style={at={(0.5,1.20)}, anchor=north, draw=black},
            enlarge x limits=0.16,
            xticklabel style={rotate=15, anchor=center, yshift=-10pt}
        ]
        
        \addplot[fill={rgb,255:red,255;green,196;blue,218}, draw=none] coordinates {(Compare positions,0.49) (Find duplicates,0.76) (Count,0.12) (Check association,0.80)};
        \addlegendentry{mistral-7b}
        
        \addplot[fill={rgb,255:red,250;green,98;blue,95}, draw=none] coordinates {(Compare positions,0.76) (Find duplicates,0.88) (Count,0.08) (Check association,0.70)};
        \addlegendentry{phi-3-small}
        
        \addplot[fill={rgb,255:red,255;green,218;blue,112}, draw=none] coordinates {(Compare positions,0.80) (Find duplicates,0.80) (Count,0.12) (Check association,0.52)};
        \addlegendentry{phi-3-medium}
        
        \addplot[fill={rgb,255:red,179;green,153;blue,212}, draw=none] coordinates {(Compare positions,0.60) (Find duplicates,0.80) (Count,0.04) (Check association,0.68)};
        \addlegendentry{gemma-2-9b}
        
        \addplot[fill={rgb,255:red,116;green,196;blue,118}, draw=none] coordinates {(Compare positions,0.75) (Find duplicates,0.80) (Count,0.16) (Check association,0.96)};
        \addlegendentry{llama-3.1-8b}

        \end{axis}
\end{tikzpicture}

%% file: tables/group.tex
\begin{table*}[t!]
    \centering
    \resizebox{0.9\textwidth}{!}{%

\begin{tabular}{lllll}
\toprule
\textbf{Model} & \textbf{Group membership} & \textbf{Group association} & \textbf{Group assoc. (alternating)} & \textbf{Iterate} \\
(Compared against) & (Sub-string search)  & (Group membership) & (Group association) & (Iterate (last)) \\
\midrule
        gpt-4-turbo & 0.96 \textcolor{green}{(0.02)}  & 0.75 \textcolor{red}{(-0.21)}  & 0.68 \textcolor{red}{(-0.07)}  & 0.83 \textcolor{red}{(-0.17)}  \\ 
    gpt-4o & 0.98 \textcolor{red}{(-0.02)} & 0.65 \textcolor{red}{(-0.33)} & 0.52 \textcolor{red}{(-0.13)} & 0.86 \textcolor{red}{(-0.14)} \\
    gpt-4o-mini & 0.96 \textcolor{red}{(-0.02)} & 0.68 \textcolor{red}{(-0.28)} & 0.52 \textcolor{red}{(-0.16)} & 0.67 \textcolor{red}{(-0.33)} \\
    cohere-command-rplus & 0.93 \textcolor{red}{(-0.07)} & 0.70 \textcolor{red}{(-0.23)} & 0.72 \textcolor{green}{(0.02)} & 0.10 \textcolor{red}{(-0.9)} \\
    mistral-7b & 0.50 \textcolor{red}{(-0.28)} & 0.57 \textcolor{green}{(0.07)} & 0.52 \textcolor{red}{(-0.05)} & 0.04 \textcolor{red}{(-0.25)} \\
    phi-3-small & 0.52 \textcolor{red}{(-0.42)} & 0.55 \textcolor{green}{(0.03)} & 0.68 \textcolor{green}{(0.13)} & 0.04 \textcolor{red}{(-0.7)} \\
    phi-3-medium & 0.60 \textcolor{red}{(-0.4)} & 0.72 \textcolor{green}{(0.12)} & 0.50 \textcolor{red}{(-0.22)} & 0.04 \textcolor{red}{(-0.69)} \\
    gemma-2-9b & 0.80 \textcolor{red}{(-0.2)} & 0.60 \textcolor{red}{(-0.2)} & 0.62 \textcolor{green}{(0.02)} & 0.14 \textcolor{red}{(-0.65)} \\
    llama-3.1-8b & 0.84 \textcolor{red}{(-0.16)} & 0.78 \textcolor{red}{(-0.06)} & 0.82 \textcolor{green}{(0.04)} & 0.05 \textcolor{red}{(-0.43)} \\
\bottomrule
\end{tabular}
}

\caption{\textbf{Results for Compute on Sets and Lists.} The numbers in parentheses indicate the performance difference compared to the corresponding tasks they are evalauted against.}
\label{tab:lists}
\end{table*}

%% file: figures/difference.tex

\begin{figure}[h]
\centering
\resizebox{0.9\columnwidth}{!}{
 \begin{tikzpicture}
        \begin{axis}[
            ybar,
            bar width=5pt,
            symbolic x coords={Compare two lists, Identify the odd group, Patch the difference},
            xtick=data,
            ymin=0, ymax=1.0,
            legend columns=4,
            legend style={at={(0.5,1.35)}, anchor=north, draw=black, font=\footnotesize},
            enlarge x limits=0.22,
            xticklabel style={rotate=10, anchor=center, yshift=-12pt},
            width=10cm, height=6cm,
        ]
        
        \addplot[fill={rgb,255:red,42;green,183;blue,202}, draw=none] coordinates {(Compare two lists,0.36) (Identify the odd group,0.78) (Patch the difference,0.37)};
        \addlegendentry{gpt-4-turbo}
        
        \addplot[fill={rgb,255:red,0;green,91;blue,150}, draw=none] coordinates {(Compare two lists,0.33) (Identify the odd group,0.87) (Patch the difference,0.89)};
        \addlegendentry{gpt-4o}
        
        \addplot[fill={rgb,255:red,173;green,216;blue,230}, draw=none] coordinates {(Compare two lists,0.29) (Identify the odd group,0.73) (Patch the difference,0.22)};
        \addlegendentry{gpt-4o-mini}
        
        \addplot[fill={rgb,255:red,254;green,138;blue,113}, draw=none] coordinates {(Compare two lists,0.32) (Identify the odd group,0.67) (Patch the difference,0.17)};
        \addlegendentry{cohere}
        
        \addplot[fill={rgb,255:red,255;green,196;blue,218}, draw=none] coordinates {(Compare two lists,0.24) (Identify the odd group,0.37) (Patch the difference,0.21)};
        \addlegendentry{mistral-7b}
        
        \addplot[fill={rgb,255:red,250;green,98;blue,95}, draw=none] coordinates {(Compare two lists,0.31) (Identify the odd group,0.52) (Patch the difference,0.55)};
        \addlegendentry{phi-3-small}
        
        \addplot[fill={rgb,255:red,255;green,218;blue,112}, draw=none] coordinates {(Compare two lists,0.30) (Identify the odd group,0.65) (Patch the difference,0.37)};
        \addlegendentry{phi-3-medium}
        
        \addplot[fill={rgb,255:red,179;green,153;blue,212}, draw=none] coordinates {(Compare two lists,0.24) (Identify the odd group,0.72) (Patch the difference,0.30)};
        \addlegendentry{gemma-2-9b}
        
        \addplot[fill={rgb,255:red,116;green,196;blue,118}, draw=none] coordinates {(Compare two lists,0.29) (Identify the odd group,0.70) (Patch the difference,0.74)};
        \addlegendentry{llama-3.1-8b}
        
        \end{axis}
    \end{tikzpicture}}
    \caption{Results for \textbf{Spot the Differences }tasks.}
    \label{fig:difference}
\end{figure}
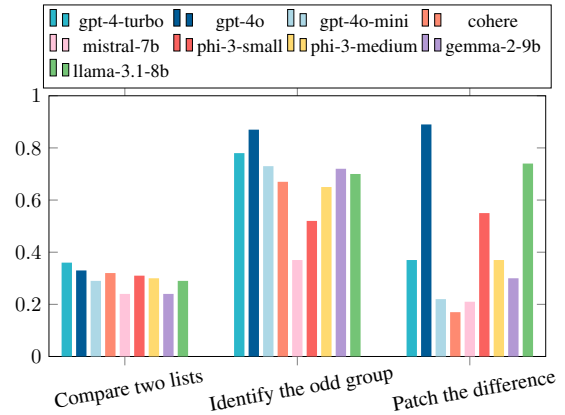

%% file: tables/state.tex
\begin{table}[t!]
\centering
    \resizebox{0.7\columnwidth}{!}{%

\begin{tabular}{lcc}
\toprule
\textbf{Model} & \textbf{Quantity state} & \textbf{Set state} \\
\midrule
gpt-4-turbo & 0.8 & \textbf{0.80} \\
gpt-4o & \textbf{1.0} & 0.65 \\
gpt-4o-mini & 0.7 & 0.24 \\
cohere & 0.0 & 0.58 \\
mistral-7b & 0.0 & 0.08 \\
phi-3-small & 0.0 & 0.13 \\
phi-3-medium & 0.0 & 0.11 \\
gemma-2-9b & 0.0 & 0.24 \\
llama-3.1-8b & 0.0 & 0.13 \\
\bottomrule
\end{tabular}
}
\caption{Results for \textbf{Stateful Processing} tasks.}
\label{tab:state}
\end{table}

%% file: figures/ablation_state_step.tex
\begin{figure}[t!]
    \centering
    \begin{subfigure}{0.49\columnwidth}
        \resizebox{\textwidth}{!}{\input{figures/ablation_quantity_state}}
    \end{subfigure}
    \begin{subfigure}{0.49\columnwidth}
        \resizebox{\textwidth}{!}{\input{figures/ablation_set_state}}
    \end{subfigure}
    \caption{Effect of context length (number of operation steps) on performance in the \textbf{quantity state} (left) and\textbf{ set state }(right) tasks.}
    \label{fig:ablation_state_step}
\end{figure}
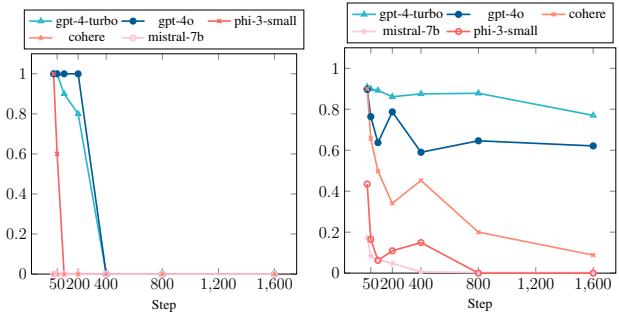

%% file: figures/ablation_quantity_state.tex
\begin{tikzpicture}
    \begin{axis}[
        xlabel={Step},
        legend style={at={(0.5,1.2)}, anchor=north, cells={align=left}, legend columns=3},
        ymin=0, ymax=1.1,
        xtick={50, 200, 400, 800, 1200, 1600},
        ytick={0,0.2,0.4,0.6,0.8,1.0},
        grid=none,
        tick label style={font=\large}
    ]

    \addplot[mark=triangle, very thick, color={rgb,255:red,42;green,183;blue,202}] coordinates {
        (25,1.0) (50,1.0) (100,0.9) (200,0.8) (400,0.0) (800,0.0) (1600,0.0)
    };
    \addlegendentry{gpt-4-turbo}

    \addplot[mark=*, very thick, color={rgb,255:red,0;green,91;blue,150}] coordinates {
        (25,1.0) (50,1.0) (100,1.0) (200,1.0) (400,0.0) (800,0.0) (1600,0.0)
    };
    \addlegendentry{gpt-4o}

    \addplot[mark=x, very thick, color={rgb,255:red,250;green,98;blue,95}] coordinates {
        (25,1.0) (50,0.6) (100,0.0) (200,0.0) (400,0.0) (800,0.0) (1600,0.0)
    };
    \addlegendentry{phi-3-small}

    \addplot[mark=star, very thick, color={rgb,255:red,254;green,138;blue,113}] coordinates {
        (25,0.0) (50,0.0) (100,0.0) (200,0.0) (400,0.0) (800,0.0) (1600,0.0)
    };
    \addlegendentry{cohere}

    \addplot[mark=o, very thick, color={rgb,255:red,255;green,196;blue,218}] coordinates {
        (25,0.0) (50,0.0) (100,0.0) (200,0.0) (400,0.0) (800,0.0) (1600,0.0)
    };
    \addlegendentry{mistral-7b}
    
    \end{axis}
\end{tikzpicture}

%% file: figures/ablation_set_state.tex
\begin{tikzpicture}
    \begin{axis}[
        xlabel={Step},
        legend style={at={(0.5,1.2)}, anchor=north, cells={align=left}, legend columns=3},
        ymin=0, ymax=1.1,
        xtick={50, 200, 400, 800, 1200, 1600},
        ytick={0,0.2,0.4,0.6,0.8,1.0},
        grid=none,
        tick label style={font=\large}
    ]

    \addplot[mark=triangle, very thick, color={rgb,255:red,42;green,183;blue,202}] coordinates {
        (25,0.909) (50,0.899) (100,0.892) (200,0.861) (400,0.875) (800,0.878) (1600,0.770)
    };
    \addlegendentry{gpt-4-turbo}

    \addplot[mark=*, very thick, color={rgb,255:red,0;green,91;blue,150}] coordinates {
        (25,0.897) (50,0.764) (100,0.637) (200,0.787) (400,0.590) (800,0.646) (1600,0.621)
    };
    \addlegendentry{gpt-4o}

    \addplot[mark=x, very thick, color={rgb,255:red,254;green,138;blue,113}] coordinates {
        (25,0.900) (50,0.657) (100,0.498) (200,0.341) (400,0.452) (800,0.200) (1600,0.088)
    };
    \addlegendentry{cohere}

    \addplot[mark=star, very thick, color={rgb,255:red,255;green,196;blue,218}] coordinates {
        (25,0.174) (50,0.084) (100,0.066) (200,0.048) (400,0.006) (800,0.003) (1600,0.003)
    };
    \addlegendentry{mistral-7b}

    \addplot[mark=o, very thick, color={rgb,255:red,250;green,98;blue,95}] coordinates {
        (25,0.435) (50,0.165) (100,0.062) (200,0.109) (400,0.149) (800,0.000) (1600,0.000)
    };
    \addlegendentry{phi-3-small}
    
    \end{axis}
\end{tikzpicture}

%% file: tables/composition.tex
\begin{table}[t!]
    \centering
        \resizebox{0.95\columnwidth}{!}{%

\begin{tabular}{lcc}
\toprule
\textbf{Model} &\textbf{ Processing Data Blocks} & \textbf{Theory of Mind }\\
\midrule
gpt-4-turbo & 0.31 &  0.26 \\
gpt-4o & \textbf{0.37} & \textbf{0.38} \\
gpt-4o-mini & 0.31 & 0.21\\
cohere & 0.26 &  0.18 \\
mistral-7b & 0.18 & 0.16\\
phi-3-small & 0.21 &  0.20\\
phi-3-medium & 0.18 &  0.10 \\
gemma-2-9b & 0.26 &  0.12 \\
llama-3.1-8b & 0.15 &  0.03 \\
\bottomrule
\end{tabular}
}
\caption{Results for the \textbf{composite tests}.}
\label{tab:comp}
\end{table}

%% file: sections/2_related_work.tex
\section{Related Work}

\paragraph{LLM Evaluation with Benchmarks} The evaluation of LLMs has traditionally relied on static benchmarks, from early benchmarks for perplexity-based  evaluation \cite{marcus-etal-1993-building} to datasets focused on specific downstream tasks such as question answering \cite{kwiatkowski2019natural}, summarization \cite{ Gliwa_2019}, math reasoning \cite{cobbe2021gsm8k}, and code generation \cite{chen2021codex}. As LLMs began to address a broader range of tasks across various domains \cite{wu2023autogenenablingnextgenllm, wang2023voyager}, more comprehensive benchmark suites \cite{hendrycks2021measuring, zhong2023agieval} were developed to assess general capabilities rather than individual task performance. Recent advancements in LLM evaluation have introduced the concept of LLM-as-a-judge, enabling the use of open-ended benchmarks without predefined answers \cite{zheng2023judging}. However, these benchmarks remain static in nature and can easily get overfit. Recently, platforms like ChatBot Arena \cite{zheng2023judging} utilize crowdsourcing to rank LLM responses and provides more dynamic evaluations. However, its reliance on human annotation makes it less scalable. Moreover, despite their utility, existing benchmarks primarily assess downstream applications that usually require multiple capabilities, making it difficult to debug and understand model weaknesses.

\paragraph{Tests and Benchmarks for Evaluating Context Utilization} As LLMs become capable of processing increasingly long inputs, designing automated tests to evaluate their ability to utilize context has become an area of active research.  A notable example is the needle-in-a-haystack (NIAH) task\footnote{\url{https://github.com/gkamradt/LLMTest_NeedleInAHaystack}}, where a small piece of information (the ``needle") is hidden within a long document, and the model needs to retrieve it. Similar tests include key-value retrieval \cite{liu-etal-2024-lost} and passkey retrieval \cite{mohtashami2023randomaccess}. The simplicity and interpretability of NIAH have made it a standard for evaluating LLM context utilization, and it has since inspired various methods for improving long-context processing \cite{mohtashami2023randomaccess, ding2024longrope, xiong2023effectivelongcontext, behrouz2024titans}.

 However, these tests focus solely on basic information retrieval, without capturing more complex aspects of context processing. To address this limitation, other tests have been proposed. Needlebench \cite{ li2024needlebench} extends simple retrieval tasks to include multi-needle reasoning and a ancestral trace task which requires navigating chains or graphs of information. \citet{song2024countingstars} introduce the Counting Stars task, which involves tallying numbers of stars embedded in phrases. Ruler \cite{hsieh2024ruler} proposes additional tasks such as variable tracking and frequent word extraction. While these tests increase task complexity or broaden the range of evaluated tasks, they remain limited in scope for systematically evaluating contextual processing.
 
Beyond individual tests, several benchmarks explicitly target long-context processing, including InftyBench \cite{zhang-etal-2024-bench}, L-Eval \cite{an-etal-2024-l}, and LongBench \cite{bai-etal-2024-longbench}. These benchmarks use NIAH-like tasks alongside question answering, summarization, and code generation over long contexts. However, like other benchmarks, they remain static and primarily measure end-to-end performance rather than systematically dissecting capabilities.

\paragraph{Analogies to Human Cognitive Testing}
Memory tests are widely used in cognitive research to assess specific functions. Such assessments often involve evaluating short term memory via recall tests \cite{crannell1957comparison, towse2008recall}, inductive reasoning via pattern recognition tasks, or attention via instruction-following\cite{kane2007identifying, nasreddine2005montreal}. By isolating distinct abilities while minimizing confounding factors like attention, memory span, and reasoning \cite{kane2007identifying}, such tests provide detailed profiles of cognitive functions, guiding interventions and shaping broader theories of human thought. Inspired by this approach, we design atomic tests that systematically isolate core aspects of LLM context processing, aiming for a fine-grained understanding of LLM memory-usage capabilities -- analogous to memory testing in humans.

%% file: sections/5_conclusion.tex
\section{Conclusions}

AI assistants powered by LLMs are expected to handle numerous operations involving memory. However, simple tests reveal that they often fall short of meeting user expectations, even in basic retrieval and processing tasks. For instance, retrieving vacation schedules for each team member from a message history that includes evolving plans over time proves challenging. To enable targeted improvements, it is essential to establish a comprehensive benchmark that tests each capability in isolation while also allowing for programmable composition to evaluate more complex scenarios. Our benchmark provides a straightforward yet effective approach to achieving this goal. We primarily focus on short-context scenarios to demonstrate that current limitations are not solely attributable to the models' challenges with parsing long contexts. Addressing these issues demands attention beyond merely solving the ``attention'' problem.

%% file: sections/999_appendix_tasks.tex
\section{Test Templates}
\label{apd:prompt_example}
In this appendix, we provide the templates of the test prompts. Placeholder context words such as ``aaa, bbb, ccc," etc., are used for illustration purposes. During testing, these context words are uniformly sampled from an English dictionary. Variable tokens in the instruction part are marked with the \textbf{bold} font.

\begin{longtable}{p{3cm}p{12cm}}
        \toprule
        \multicolumn{2}{c}{\textbf{Search} } \\
        \midrule
         Task name & Prompt  \\
         \midrule
         String search (with word) & \textit{Context:}
         
         aaa, bbb, ccc, ...
         
         \textit{Instruction:}
         
         Given the context, determine if the word ``\textbf{bbb}" is present in the context. Answer with ``yes'' or `no''.
         
         Answer: \\
         
         \midrule
         String search (with subsequence) & \textit{Context:}
         
         aaa, bbb, ccc, ...
         
         \textit{Instruction:}
         
         Given the list of words in the context, determine if the sequence ``\textbf{bbb, xxx, ddd}'' appears in the context. Answer with `yes' or 'no'.
         
         Answer: \\
         
         \midrule
         
         Key-value search & \textit{Context:} 
         
         aaa:bbb, ccc:ddd, ...
         
         \textit{Instruction:}
         
         Given a list of word pairs formatted as ``word\_1: word\_2'' in the context, return the second word associated with the provided first word. For the first word ``\textbf{aaa}'', the corresponding second word is:\\
         \midrule
         Batch search & \textit{Context: }
         
         aaa:bbb, ccc:ddd, ...
         
        \textit{Instruction:}
         
         Given a list of word pairs formatted as ``word\_1: word\_2'' in the context, return the second word associated with the provided first words. For the first words: \textbf{aaa, ccc, ...}, the corresponding second words are:\\
        \bottomrule
        \toprule
        
        \multicolumn{2}{c}{\textbf{Recall and Edit} } \\
        \midrule
         Task name & Prompt  \\
        \midrule
        Snapshot & \textit{Context:}
        
        aaa, bbb, ccc, ...
        
        \textit{Instruction:}
        
        Repeat the previous context exactly as it is, without making any additions or deletions.
        
        Answer:\\
        \midrule

        Replace all (x to y) & \textit{Context:}

        aaa, bbb, aaa, ccc, aaa, ddd, ...

        \textit{Instruction:}
        
        Repeat the previous context and replace the word ``\textbf{aaa}'' with ``\textbf{zzz}'' each time it appears.
        
        Answer: \\
        \midrule
        Replace all (x to null) & \textit{Context:}

        aaa, bbb, aaa, ccc, aaa, ddd, ...

        \textit{Instruction:}
        
        Repeat the previous context but skip the word ``\textbf{aaa}'' each time it appears.
        
        Answer: \\
        \midrule
        Overwrite positions (nth to y) & \textit{Context:}

        aaa, bbb, ccc, ...
        
        \textit{Instruction}:
        
        Repeat the previous context and replace every \textbf{third} word with ``zzz''.
        
        Answer: \\
        \midrule
        Overwrite positions (nth to null) & \textit{Context:}

        aaa, bbb, ccc, ...
        
        \textit{Instruction}:
        
        Repeat the previous context and skip every \textbf{other} word.
        
        Answer: \\
        \midrule

        Functional updates & \textit{Context:}

        111, 222, 333, ...

        \textit{Instruction:}
        
        \textbf{Add 3} to every number in the previous context.
        
        Answer: \\
        \bottomrule
        \toprule
        \multicolumn{2}{c}{\textbf{Match and Compare} } \\
        \midrule
         Task name & Prompt  \\
        \midrule

        Compare positions & \textit{Context}:
        
        aaa, bbb, ccc, ...
        
        \textit{Instruction}:
        
        Given the list of words in the context, determine the relative positions of two words. Does the word ``\textbf{aaa}'' appear before the word ``\textbf{ccc}'' in the list? Answer ``yes'' or ``no''.
        
        Answer: \\

        \midrule

        Find duplicates & \textit{Context}:
        
        aaa, bbb, aaa, ...
        
        \textit{Instruction}:
        
        A word is repeated multiple times in the context. Your task is to identify the word that is repeated.
        
        The repeated word is: \\

        \midrule

        Count & \textit{Context}:
        
        aaa, bbb, aaa, ...
        
        \textit{Instruction}: 
        
        Count the number of times the word ``\textbf{aaa}'' appeared in the context.
        
        Answer: The word ``\textbf{aaa}'' appeared \\

         \midrule

        Check association & \textit{Context}:
               
        aaa:attribute 1, bbb:attribute 2, ccc: attribute 2, ddd: attribute 1, ...

        \textit{Instruction}:

        Given the list of words and their respective attributes in the format of ``word:attribute'', determine if the word ``\textbf{aaa}'' and the word ``\textbf{ggg}'' have the same attribute. Answer with ``yes'' or ``no''.

        Answer: \\

        \bottomrule
        \toprule
        \multicolumn{2}{c}{\textbf{Spot the Differences} } \\
        \midrule
         Task name & Prompt  \\
        \midrule

        Compare two lists & \textit{Context}:
        
        List 1: aaa, bbb, ccc, ...
        
        List 2: aaa, ddd, ccc, ...
        
        \textit{Instruction}: 
        
        There are two lists of words in the context. The first list contains the original words. The second list is similar to the first but has some words replaced with different ones. Your task is to identify the words in the \textbf{first/second} list that are different from those in the other list. Provide the different words as your answer.
        
        Answer: \\

        \midrule

        Identify the odd group & \textit{Context}:
        
        List 1: aaa, bbb, ccc, ...
        
        List 2: bbb, aaa, ccc, ...
        
        List 3: aaa, zzz, ccc, ...
        
        List 4: ccc, aaa, bbb, ...
        
        \textit{Instruction}:
        
        Given the lists of words in the context, identify the list that is different from the others. Provide the list number as your answer. For example, if the Nth list is different, provide ``List N" as your answer.
        
        Answer: \\
        \midrule
        Patch the difference & \textit{Context}:
        
        aaa, bbb, ccc, aaa, bbb, ccc, ...
        
        \textit{Instruction}: 
        
        Given the sequence of words that follows a specific pattern in the context, predict the \textbf{Nth} word that appears after the final word in the given sequence.
        
        Answer: The \textbf{Nth} word that appears after the final word in the given sequence is\\

        \bottomrule
        \toprule

        \multicolumn{2}{c}{\textbf{Compute on Sets and Lists} } \\
        \midrule
         Task name & Prompt  \\
        \midrule

        Group membership & \textit{Context}:
        
        List 1: aaa, bbb, ccc, ...\newline
        List 2: ddd, eee, fff, ...\newline
        ...
        
        \textit{Instruction}:
        
        Given the lists of words in the context, determine which list contains the word ``\textbf{fff}". If the word is not present in either list, answer ``no".
        
        Answer:\\

        \midrule
        Group association & \textit{Context}:
        
        List 1: aaa, bbb, ccc, ...\newline
        List 2: ddd, eee, fff, ...\newline
        ...
        
        \textit{Instruction}:
        
        Given the lists of words in the context, determine if the word ``\textbf{aaa}'' and the word ``\textbf{eee}'' are in the same list. Answer with ``yes'' or ``no''.

        Answer: \\
        \midrule

        Group association (alternating) & Context:

        Role A: aaa, bbb, ...\newline
        Role B: ccc, ddd, ...\newline
        Role A: eee, fff, ...\newline
        Role B: ggg, hhh, ...\newline
        ...

        Instruction:

        Given the context with alternating roles and their respective context words, determine if the word ``\textbf{aaa}'' and the word ``\textbf{ggg}'' are in the same role. Answer with ``yes'' or ``no''.

        Answer:\\

        \midrule
        Iterate & \textit{Context}:
        
        List 1: aaa, bbb, ccc, ...\newline
        List 2: ddd, eee, fff, ...\newline
        ...
        
        \textit{Instruction}:
        
        Given the lists of words in the context, identify and recall the \textbf{last} word from each list. Provide your answer as a list of these words separated by commas. \\
        & Answer:
        \\
 
        \bottomrule
        \toprule

        \multicolumn{2}{c}{\textbf{Stateful Processing} } \\
        \midrule
         Task name & Prompt  \\

        \midrule
        Quantity state & \textit{Context}:
        
        Begin with the number xx. Perform the following operations:\newline
        1. Add xx
        2. Subtract xx
        3. ...
    
        \textit{Instruction}:
        
        In the context, you are given an initial number and a series of operations to perform on that number. Your task is to determine the final result of the operations. Write your final answer after the text ``FINAL ANSWER:". For example, ``FINAL ANSWER: 42".
        
        FINAL ANSWER:\\ 
        \midrule
        Set state & \textit{Agent actions}:
        
        Agent draws aaa, bbb, ccc\newline
        Agent discards bbb, ccc\newline
        Agent draws ddd, fff\newline
        Agent discards ddd\newline
        ...

        \textit{Instruction}:
        
        Given the actions of the agent, your task is to determine the final list of words the agent ends up with after a series of actions. Write your final answer after the text ``FINAL ANSWER:". For example, ``FINAL ANSWER: word1, word2, word3".
        
        FINAL ANSWER:\\
       
        \bottomrule

        \toprule
        \multicolumn{2}{l}{\textbf{Processing Data Blocks} } \\
        \midrule
         Task name & Prompt  \\
        \midrule
        Processing Data Blocks & \textit{Context}:
        
        Role 1: aaa, bbb, ccc, ...\newline
        Role 2: ddd, eee, fff, ...\newline
        Role 3: ggg, hhh, iii, ... \newline
        Role 1: jjj, kkk, ... \newline
        ...
        
        \textit{Instruction}:
        
        The context consists of a series of alternating roles, each associated with a list of words. Your task is to identify and recall all the words from the role labeled ``{\textbf{Role 2}}" that appear after the word ``{\textbf{zzz}}" in the sequence. Please write your answer after the text ``Answer:". For example, ``Answer: word1, word2, word3".
        
        Answer:\\
        \bottomrule
        \toprule
        \multicolumn{2}{c}{\textbf{Composite-State Tracking (Theory of Mind)} } \\
        \midrule
         Task name & Prompt  \\
        \midrule
        Theory of Mind & \textit{Agents actions:}:
        
        Agent A starts with the following words: aaa, bbb, ccc, ...\newline
        Agent B starts with the following words: ddd, eee, fff, ...\newline
        Agent B starts with the following words: ggg, hhh, iii, ...\newline
        Agent B swaps the following words ``ddd" with Agent C for the following words ``hhh". \newline
        Agent A discards the following words: ccc, zzz, .... \newline
        Agent C draws the following words: xxx, yyy, .... \newline
        ...
        
        \textit{Instruction}:
        
        Given the actions of the agents, your task is to determine the final list of words each agent ends up with after a series of actions. Write your final answer after the text ``FINAL ANSWER:". For example, ``FINAL ANSWER: Agent A: word1, word2, word3\textbackslash nAgent B: word4, word5".
        
        FINAL ANSWER:\\
        \bottomrule
        
    \label{tab:level1}
\end{longtable}

%% file: sections/999_appendix_hyperparams.tex
\section{Task Details}
\label{apd:task_detail}

This appendix provides details for each task, including the number of examples, evaluation metrics, and configurable hyperparameters. The context length is fixed at 4k for almost all tasks, apart from Stateful Processing, where the context is determined by number of operation steps and set to 200 for quantity state and 100 for set state, which maps to around 1.5k context tokens.

Here is an example of number of examples calculation String search (with word):  5 (query depth) *  2 (labels)  * 5 (samples per parameter setting) = 50.

\begin{longtable}{lp{6.5cm}cc}
\caption{Task Overview with Hyperparameters, Number of Examples, and Evaluation Metrics\label{tab:task_details}} \\

\toprule
\textbf{Task Name} & \textbf{Hyperparameters} & \textbf{\# of Examples} & \textbf{Metric} \\ 
\midrule
\endfirsthead

\toprule
\textbf{Task Name} & \textbf{Hyperparameters} & \textbf{\# of Examples} & \textbf{Metric} \\ 
\midrule
\endhead

\midrule
\multicolumn{4}{r}{\textbf{Continued on next page}} \\ 
\midrule
\endfoot

\bottomrule
\endlastfoot

\multicolumn{4}{l}{\textbf{Search}} \\ 
String search (word) & query depth = [0, 0.25, 0.5, 0.75, 1], label = [positive, negative], samples = 5 & 50 & exact\_match \\
String search (sequence) & sequence length = [8, 16, 32, 64], label = [positive, negative], samples = 10 & 80 & exact\_match \\ 
Key-value search & query depth = [0, 0.25, 0.5, 0.75, 1], samples = 10 & 50 & exact\_match \\ 
Batch search & batch size = [4, 8, 16, 32], samples = 5 & 20 & rouge-L\_recall \\
\midrule
\multicolumn{4}{r}{\textbf{Number of Entries for Category: 200}} \\ 
\midrule

\multicolumn{4}{l}{\textbf{Recall and Edit}} \\ 
Snapshot (words) & samples = 10 & 10 & rouge-L \\
Replace all & density = [0.2, 0.4, 0.6, 0.8], y = [random word, null], samples = 5 & 40 & rouge-L \\
Overwrite positions & nth = [2, 3, 4], y = [random word, null], samples = 5 & 30 & rouge-L \\ 
Snapshot (numbers) & samples = 10 & 10 & rouge-L \\ 
Functional updates & function type = [add (3), subtract (1), multiply (2)], samples = 5 & 15 & rouge-L \\ 
\midrule
\multicolumn{4}{r}{\textbf{Number of Entries for Category: 105}} \\ 
\midrule

\multicolumn{4}{l}{\textbf{Match and Compare}} \\ 
Compare positions & query 1 depth = [0, 0.25, 0.5, 0.75, 1], query 2 depth = [0, 0.25, 0.5, 0.75, 1], samples = 3 & 75 & exact\_match \\ 
Find duplicates & repetition = [2, 4, 8, 16, 32], samples = 5 & 25 & exact\_match \\ 
Count & repetition = [2, 4, 8, 16, 32], samples = 5 & 25 & exact\_match \\ 
Check association & n attribute = [2, 4, 8, 16, 32], label = [positive, negative], samples = 5 & 50 & exact\_match \\
\midrule
\multicolumn{4}{r}{\textbf{Number of Entries for Category: 175}} \\ 
\midrule

\multicolumn{4}{l}{\textbf{Spot the Differences}} \\ 
Compare two lists & num different words = [1, 5, 10, 20], chosen list = [first, second], samples = 10 & 80 & rouge-L\_recall \\ 
Identify the odd group & words per group = [25, 50, 75, 100], percentage of difference = [0, 0.25, 0.5], samples = 5 & 60 & exact\_match \\ 
Patch the difference & pattern length = [2, 15, 30],  cut off depth = [0, 0.5, 1], nth = [1, 3, 6], samples = 5 & 120\footnote{For \textit{Patch the difference} task with pattern length 2, there is only two cut off percentage options; therefore the total number of data points is 120 instead of 135.} & exact\_match \\ 
\midrule
\multicolumn{4}{r}{\textbf{Number of Entries for Category: 260}} \\ 
\midrule

\multicolumn{4}{l}{\textbf{Compute on Sets and Lists}} \\ 
Group membership & number of groups = [4, 8, 16, 32], query depth = [0, 0.25, 0.5, 0.75, 1], samples = 5 & 100 & exact\_match \\ 
Group association & number of groups = [4, 8, 16, 32], label = [positive, negative], samples = 5  & 40 & exact\_match \\ 
Group association (alternating) & number of groups = [2, 4, 8, 16, 32], number of turns = 10, label = [positive, negative], sample = 5 & 50 & exact\_match \\
Iterate & number of groups = [4, 8, 16, 32], samples = 5 & 20 & rouge-L \\ 
\midrule
\multicolumn{4}{r}{\textbf{Number of Entries for Category: 210}} \\ 
\midrule

\multicolumn{4}{l}{\textbf{Stateful Processing}} \\ 
Set state & number of steps = 100, set size = [5, 10, 15, 20], samples = 10 & 40 & jaccard\_similarity \\ 
Quantity state & number of steps = 200, samples = 10 & 10 & exact\_match \\ 
\midrule
\multicolumn{4}{r}{\textbf{Number of Entries for Category: 50}} \\ 
\midrule

\multicolumn{4}{l}{\textbf{Composite}} \\ 
Processing data blocks & number of blocks = [2, 4, 8, 16, 32], number of turns = 10, samples = 5 & 50 & rouge-L \\
Theory of mind & number of steps = 100, number of agents = [2, 3, 4], samples = [10, 20] & 60 & jaccard\_similarity \\ 
\midrule
\multicolumn{4}{r}{\textbf{Number of Entries for Category: 110}} \\ 
\midrule

\multicolumn{4}{r}{\textbf{Total Number of Entries: 1110}} \\ 
\end{longtable}

\section{Evaluation Metrics}
\label{apd:eval}
In this appendix section, we provide details about the evaluation metrics we have used in the tests.

\begin{itemize} \item Exact Match: The exact match accuracy measures whether the generated answer exactly matches the reference answer. It is computed as follows: 
\[
\text{Exact Match} = 
\begin{cases} 
1 & \text{if } \text{reference\_answer} = \text{generated\_answer}, \\
0 & \text{otherwise}.
\end{cases}
\]
\item ROUGE-L / ROUGE-L-recall: ROUGE (Recall-Oriented Understudy for Gisting Evaluation) \cite{lin-2004-rouge} measures the verbatim overlap between the reference and the generated answers. ROUGE-L specifically looks for the longest common subsequence (LCS) between the two texts, which reflects the structure of the text and the longest sequence of matching words. ROUGE-L recall focuses on the ability of the model to recall the content from the reference answer, and it emphasizes matching the longest subsequences.

ROUGE-L-recall can be defined as:
\[
\text{ROUGE-L-recall} = \frac{LCS(\text{generated\_answer}, \text{reference\_answer})}{\text{length of reference\_answer}}
\]

ROUGE-L is computed as the F1-score, which combines both precision and recall to provide a more balanced measure of overlap.

\item Jaccard Similarity: Jaccard similarity measures the overlap between two sets by comparing the intersection and union of the sets. It is computed as:
\[
\text{Jaccard Similarity} = \frac{|A \cap B|}{|A \cup B|}
\]
where \(A\) and \(B\) are sets representing the elements in the generated and reference answers, respectively. This metric is used for tasks involving set-based comparisons or when the goal is to measure the similarity between two sets of elements (e.g., word sets).
\end{itemize}

\section{More Examples for Context Length Variation and Prompt Variation}

\label{apd:more_examples}
In this section, we present additional examples illustrating the effects of context length and minor variations in prompt instructions on selected models.

\paragraph{Length variation} Table~\ref{tab:added_length_variation} reports results from three additional tasks: \textit{string search (word)}, \textit{replace all}, and \textit{iterate}. The results reaffirm that while models tend to perform well on the search task even at long context lengths, their performance on other tasks degrades significantly at much shorter lengths. This highlights a fundamental limitation in how effectively these models utilize long-range context for non-search tasks.

\begin{table}[h]
\centering
\begin{subtable}[t]{0.48\textwidth}
\centering
\begin{tabular}{llllll}
\toprule
Model       & 2000 & 4000 & 8000 & 16000 & 32000 \\
\midrule
gpt-4o      & 1.00 & 1.00 & 1.00 & 1.00  & 0.98  \\
gpt-4o-mini & 0.98 & 0.98 & 0.94 & 0.90  & 0.78  \\
phi-3-small & 1.00 & 0.94 & 0.90 & 0.98  & 0.98  \\
\bottomrule
\end{tabular}
\caption{String search (word)}
\end{subtable}
\hfill
\begin{subtable}[t]{0.48\textwidth}
\centering
\begin{tabular}{llllll}
\toprule
Model       & 1000 & 2000 & 4000 & 8000 & 16000 \\
\midrule
gpt-4o      & 1.00 & 1.00 & 0.99 & 0.81 & 0.48  \\
gpt-4o-mini & 0.99 & 0.91 & 0.84 & 0.71 & 0.42  \\
phi-3-small & 0.87 & 0.67 & 0.49 & 0.32 & 0.06  \\
\bottomrule
\end{tabular}
\caption{Replace all}
\end{subtable}

\vspace{0.5em}

\begin{subtable}[t]{0.48\textwidth}
\centering
\begin{tabular}{llllll}
\toprule
Model       & 1000 & 2000 & 4000 & 8000 & 16000 \\
\midrule
gpt-4o      & 1.00 & 0.97 & 0.86 & 0.70 & 0.57  \\
gpt-4o-mini & 0.91 & 0.87 & 0.67 & 0.43 & 0.28  \\
phi-3-small & 0.15 & 0.09 & 0.04 & 0.01 & 0.01  \\
\bottomrule
\end{tabular}
\caption{Iterate}
\end{subtable}
\caption{Model performance across varying context lengths for three tasks.}
\label{tab:added_length_variation}
\end{table}

\paragraph{Prompt variation.} We provide additional results examining how small changes in prompt phrasing affect model performance across three tasks (see Table \ref{tab:prompt_variation_additional}). In general, we find that minor wording changes (e.g., in \textit{replace all} and \textit{quantity state}) do not significantly affect performance. This suggests that task accuracy is primarily driven by the model's underlying capability to process memory rather than sensitivity to prompt wording. However, more substantial prompt changes, such as shown in \textit{check association} task, can lead to notable differences in performance across models. To minimize the impact of prompt tuning, we standardize all prompts to the versions specified in the main paper.

\begin{table*}[h]
\centering
\begin{subtable}[t]{0.48\textwidth}
\centering
\begin{tabular}{lccccc}
\toprule
Model & Var 1 & CI$_{95\%}$ & Var 2 & CI$_{95\%}$ \\
\midrule
gpt-4o        & 0.99 & (0.93, 1.00) & 0.98 & (0.93, 1.00) \\
gpt-4o-mini   & 0.84 & (0.71, 0.94) & 0.83 & (0.71, 0.94) \\
phi-3-small   & 0.49 & (0.32, 0.63) & 0.51 & (0.35, 0.65) \\
\bottomrule
\end{tabular}
\caption{\textbf{Replace all} (ROUGE-L)}
\end{subtable}
\hfill
\begin{subtable}[t]{0.48\textwidth}
\centering
\begin{tabular}{lccccc}
\toprule
Model & Var 1 & CI$_{95\%}$ & Var 2 & CI$_{95\%}$ \\
\midrule
gpt-4o        & 0.72 & (0.58, 0.83) & 0.76 & (0.63, 0.86) \\
gpt-4o-mini   & 0.60 & (0.46, 0.72) & 0.58 &  (0.42, 0.69) \\
phi-3-small   & 0.70 & (0.56, 0.81) & 0.60 & (0.46, 0.72) \\
\bottomrule
\end{tabular}
\caption{\textbf{Check association} (Exact match)}
\end{subtable}

\vspace{0.5em}

\begin{subtable}[t]{0.48\textwidth}
\centering
\begin{tabular}{lccccc}
\toprule
Model & Var 1 & CI$_{95\%}$ & Var 2 & CI$_{95\%}$ \\
\midrule
gpt-4o        & 1.00 & (0.72, 1.00) & 1.00 & (0.72, 1.00) \\
gpt-4o-mini   & 0.70 & (0.40, 0.89) & 0.80 & (0.49, 0.94) \\
phi-3-small   & 0.00 & (0.00) & 0.00 & (0.00) \\
\bottomrule
\end{tabular}
\caption{\textbf{Quantity state} (Exact match)}
\end{subtable}

\caption{Effect of prompt variations on model performance across three tasks. We report accuracy (or ROUGE-L) and 95\% confidence intervals.}
\label{tab:prompt_variation_additional}
\end{table*}

\noindent\textbf{Prompt variants used in the table above:}
\begin{itemize}
    \item \textbf{Replace all}
    \begin{itemize}
        \item Var 1: ``Repeat the previous context and replace the word \texttt{aaa} with \texttt{bbb}.''
        \item Var 2: ``Copy the previous context but replace the word \texttt{aaa} with \texttt{bbb}.''
    \end{itemize}
    
    \item \textbf{Check association}
    \begin{itemize}
        \item Var 1: ``Given the context with words and their assigned attributes in the format of \texttt{word: ATT\_N}, determine if the word \texttt{aaa} has the same attribute as the word \texttt{bbb}?''
        \item Var 2: ``Given the context''
    \end{itemize}
    
    \item \textbf{Quantity state}
    \begin{itemize}
        \item Var 1: ``Your task is to determine the final result of the operations.''
        \item Var 2: ``Determine the final result after the operations.''
    \end{itemize}
\end{itemize}